\documentclass[11pt]{article}

\usepackage[preprint]{acl}

\usepackage{times}
\usepackage{latexsym}
\usepackage[T1]{fontenc}
\usepackage[utf8]{inputenc}
\usepackage{microtype}
\usepackage{inconsolata}
\usepackage{graphicx}
\usepackage{subcaption}
\usepackage{booktabs}
\usepackage{amsmath}
\usepackage{amssymb}
\usepackage{nicefrac}
\usepackage{enumitem}
\usepackage{placeins}

\title{Language Models as Measurement Apparatus for Culture}

\author{Kent K. Chang \\
  School of Information \\
  University of California, Berkeley \\
  \texttt{kentkchang@berkeley.edu}}

\begin{document}
\maketitle

\begin{abstract}
Language models are increasingly used to quantify cultural phenomena, but what makes such measurement distinctively \textit{cultural}?
This paper argues that NLP work on culture is a \textit{material-discursive practice}: the apparatus---model, data, annotation, evaluation---participates in constituting the cultural reality it measures, rather than passively recording it.
Drawing on Karen Barad's concept of the \textit{agential cut}---the contingent boundary between phenomenon and instrument---I show that the apparatus's substantive design choices draw such boundaries, and that the boundary is entangled from the start because language models have already internalized much of the cultural material they measure.
I illustrate this through three case studies on television and film dialogue (measuring structure, interaction, and deviation) and three examinations of the apparatus itself (erasure of cultural markers, attunement to historical material, and agency in an agentic workflow).
This big picture analysis proposes a research program that is theory-driven, empirically rigorous, and culturally contingent, treating each agential cut as a conscious commitment, at once methodological and ethical.\footnote{Accepted to the Big Picture workshop co-located with ACL 2026. This version expands the camera-ready in \textit{Proceedings of The Big Picture v2: Crafting a Research Narrative}, pp. 131--143, San Diego, CA, USA. Association for Computational Linguistics.}
\end{abstract}

\section{Introduction}\label{sec:intro}

A growing body of NLP research engages with cultural objects, often under the rubric of cultural analytics: literary texts, social media, and other artifacts whose significance is irreducible to information content~\citep{Piper2016-hr,piper2017} and constitutes a symbolic form~\citep{Cassirer1923-yh}.
For instance, word embeddings have traced historical shifts in cultural concepts~\citep{garg2018word,hamilton-etal-2016-diachronic}, which has been shown to be robust for humanistic inquiries~\cite{zhou-etal-2025-rethinking}, and their contextualized variants have mapped the geometry of social meaning~\citep{kozlowski2019geometry,lucy2022discovering}. Connotation frames have measured implicit power and agency in film dialogue~\citep{sap2017connotation,antoniak-etal-2023-riveter}; computational sociolinguistics has modeled stylistic coordination in dialogue~\citep{Danescu-Niculescu-Mizil2011-dz}; and large language models (LLMs) have been probed for cultural knowledge~\citep{chiu2025culturalbench}.
This work leverages the affordance of NLP methods to address cultural questions, attending to both empirical rigor and interpretive depth such methods enable.
What remains incomplete in the bigger picture is an explicit account of what it means to \textit{measure} culture---as opposed to measuring sentiment, syntax, or factual accuracy.

Recent work has begun to address this gap: \citet{zhou-etal-2025-culture} deftly draws on sociocultural linguistics to argue that cultural NLP needs a coherent theory of culture grounded in indexicality, positionality, and emergence.
Building on this, this paper offers a big picture analysis to expound on what it means to \textit{measure} culture with language models, asking: What happens when a language model is used as an instrument of cultural measurement?
I argue that this work constitutes a \textit{material-discursive practice}~\citep{barad2007,brown2000social}: the material configuration of the apparatus (model architecture, training data) and the discursive framework of the researcher (annotation categories, evaluation criteria, interpretive commitments) are entangled in the measurement and inseparable from it.

To make this concrete, I develop the concept of the~\textit{agential cut}~\citep{barad2007} for cultural analytics: the contingent boundary that an apparatus enacts between what counts as phenomenon and what counts as instrument.
In using computational methods to study culture, every design choice---model architectures, taxonomies for classification, adjudication of annotation---draws such a boundary, and the boundary could always have been drawn differently.
At the same time, what makes language models distinctive when applied to cultural artifacts is that they often have already encountered snippets~\citep{chang-etal-2023-speak} or summaries of the cultural material they measure during pre-training: the boundary between instrument and object is therefore inevitably entangled from the start.
As LLMs are increasingly used for social and cultural measurements~\citep{Bamman2024-yg,Halterman2025-iu}, this entanglement raises a problem that is prior to, and distinct from, the representational gap between data and cultural reality~\citep{Bode2020-fr}: while data remains a partial construction of the world, the LLM (or any measuring instrument) has already internalized the very material it is asked to measure.
The case studies in this paper return to this entanglement repeatedly. 

I develop the argument through three case studies on dialogic interactions found in film and television, a site where social identities are constructed and contested, with a rich tradition spanning film and media studies~\citep{Kozloff2000-pe}, sociolinguistics~\citep{Richardson2010-kg,Bednarek2023-zu}, and conversational pragmatics~\citep{Lakoff1984-sd,He1998-rl}.
Each case study represents a type of cultural measurement: \textit{structure} (conversation disentanglement), \textit{interaction} (role attribution and gender), and \textit{deviation} (stereotypic relation extraction).
The measurements produced---gendered patterns in conversational agency, disparities in role attribution, the formalization of subversion---are constitutively contingent.

With those examples of cultural measurements, I turn to the inextricable relationship between measuring instruments and cultural artifacts: What if we mask out---or, \textit{erase}---certain cultural markers to perturb the apparatus? 
What happens when we adapt---or, \textit{attune}---the underlying model to not just the task, but also a specific historical period mediated by language?
What if we distribute the measurement across a workflow of autonomous models (or, \textit{agents}), each carrying its own memory of the culture it measures?

\section{Measurements of culture}\label{sec:framework}

\begin{table*}[t]
\centering
\small
\begin{tabular}{p{1.5cm}p{4.2cm}p{4.0cm}p{3.8cm}}
\toprule
\textbf{Type} & \textbf{Apparatus configuration} & \textbf{Cultural question} & \textbf{Source of contingency} \\
\midrule
Structure & Multi-party dialogue\newline \hspace*{0.5em}$\to$ directed reply-to graph\newline \hspace*{0.5em}\citep{chang-etal-2023-dramatic} & How does conversational agency emerge from the \textbf{topology of who-responds-to-whom}? & architecture and input representation define what counts as a thread \\
\addlinespace
Interaction & Audio-visual signal\newline \hspace*{0.5em}$\to$ Goffmanian role labels\newline \hspace*{0.5em}\citep{chang2026multimodalconversationstructureunderstanding} & How are speaking and listening roles \textbf{distributed across participants}? & category taxonomy and modality selection determine which roles are measurable \\
\addlinespace
Deviation & Dyadic dialogue\newline \hspace*{0.5em}$\to$ stereotypic relationship type\newline \hspace*{0.5em}\citep{chang2024subversive} & Where does performed interaction \textbf{depart from normative expectation}? & norm is learned from training; deviation exists only relative to a trained baseline \\
\bottomrule
\end{tabular}
\caption{Three modes of cultural measurement, organized by the type of agential cut the apparatus enacts.}
\label{tab:taxonomy}
\end{table*}

\subsection{Operationalization and measurement}\label{sec:operationalization}

The dominant framework for computational work on culture derives from the social sciences, where \textit{operationalization}---translating a theoretical concept into a measurable variable---sits at its core.
\citet{moretti2013} influentially imported this into digital humanities, and subsequent work has refined the call for the computational study of culture to be explicit about its methodological commitments~\citep{piper2017,underwood2019}.
Two recent positions extend this concern to AI:
\citet{pmlr-v267-wallach25a} argue that evaluating generative AI systems is fundamentally a measurement challenge where constructs such as helpfulness or harm cannot be treated as natural labels but must be defined, instrumented, and validated.
\citet{Kommers2025-ol} reframe the problem from metric selection to representational format, arguing that ``thin'' numerical descriptions strip away the cultural context that gives human activity meaning, and that LLMs might instead scale the ``thick description'' \citep{Geertz1973-bj} developed in the interpretive humanities.

Measurements of culture, in the sense developed below, take up the same problem from yet a third angle. The question is not merely one of how we measure AI systems, but also how AI systems measure cultural artifacts.
Plays, novels, films, and television scenes are not raw observations of social variables; they are organized by genre convention, historical context, audience reception, and critical interpretation.
Computational methods can make such artifacts measurable at scale, but only by making them legible in some particular way: as character lines, reply graphs, role labels, relationship types, or deviations from a learned norm.

\begin{figure}[t]
  \centering
  \includegraphics[width=\columnwidth]{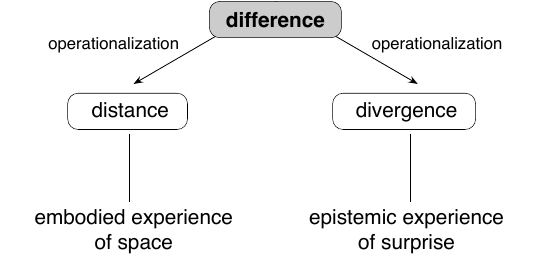}
  \caption{Two paths to operationalizing the concept of difference in computational work on culture~\citep{chang2020divergence}: distance (a spatial metaphor, subject to metric axioms including symmetry) and divergence (a cognitive metaphor, capturing asymmetric relationships invisible to distance). The choice between them is itself an agential cut that determines which cultural relationships become measurable.}
  \label{fig:divergence-tree}
\end{figure}

As \citet{piper2017} argues in his framework for literary modeling, operationalization is itself a reduction---and such reduction is where the measurement happens.
Consider a simple example of operationalization---measuring how much two texts differ. Distance metrics like cosine afford spatial proximity arguments and inherit the symmetry of a metric space; divergences like Kullback--Leibler afford asymmetric arguments about encoding cost and surprise (Fig.~\ref{fig:divergence-tree}; \citealp{chang2020divergence}).
The two operationalize the same word---\textit{difference}---into different cultural facts.
The choice between them determines which cultural relationships are made measurable.

Operationalization is productive, but it carries an assumption: that the concept being operationalized exists independently of the measurement procedure.
Recent work in cultural analytics questions this: 
\citet{mcnulty2025computation} reconsider the relationship between computation and form in the wake of generative AI, treating models themselves---their architectures and outputs---as cultural-technical ``forms'' that both enable and require new modes of analysis.
\citet{dobson2025beyond} argues that architecture is not a neutral container but a substantive interpretive choice---a site where meaning is made and historicity registered.
These observations suggest a framework that treats the entire configuration---model, data, annotation, evaluation---as a larger, coherent whole: indeed, an \textit{apparatus} that participates in producing its object, not one treating data, task, and algorithm as separable components of a linear research narrative.

\subsection{Entanglement and cuts}\label{sec:apparatus}

The linear research narrative---task, data, model, metric---works for many NLP problems by treating these components as separable stages.
For cultural measurement, that separation collapses from both sides.
Decisions about what to model---how to represent a fictional character, what counts as an interaction, which categories to annotate, how much context to expose, for example---are not preliminary to the measurement but constitutive of it; they determine which cultural realities can emerge and which are subsequently foreclosed.

Crucially, the instrument itself is culturally formed: what an LLM knows about culture is what circulates, and what circulates is structured by prestige and the cultural industries long before any researcher uses the model to measure.
The problem is not that the model has this cultural past---for measurements of culture, some past is often necessary---but whether that past is acknowledged, tested, and interpreted as part of the apparatus.
This is related to, but distinct from, the ontological gap between data and cultural reality that \citet{Bode2020-fr} identifies.

In machine learning terms, decisions about what to model are usually treated as task design, and the cultural formation of the instrument as contamination or memorization~\citep{mallen-etal-2023-trust} that puts pressure on model reliability.
Those terms are useful, but too narrow: they treat the apparatus and its object as separable when, for cultural measurement, they are not.
From \citeposs{barad2007} reading of Niels Bohr, I take the concept of the \textit{agential cut}: the boundary an apparatus enacts between what counts as phenomenon and what counts as instrument.
I use the term here in a constrained methodological sense---not every implementation detail is an agential cut.
Random seeds, batch sizes, choice of GPU vendor, logging verbosity: these do not, in any normal range, change what cultural phenomenon can appear in the measurement.
A design choice becomes a cut when it does: the label taxonomy, the context window, the modality, the anonymization procedure, the training data, or the norm against which deviation is measured.

The case studies that follow each enact a different cut (summarized in~Table~\ref{tab:taxonomy}).
Conversation disentanglement cuts continuous dialogue into a directed reply-to graph~\citep{chang-etal-2023-dramatic}.
Conversational role attribution cuts an audio-visual scene into speaker, addressee, and side-participant roles~\citep{chang2026multimodalconversationstructureunderstanding}.
Stereotypic relation extraction cuts a trained expectation into a norm against which performance can depart~\citep{chang2024subversive}.
Each measurement is empirical, but none is independent of the apparatus that produces it.
Framed like this, cultural analytics seeks to hold together the positivist work (of building models that shed light on culture) with the critical work (of insisting on the contingency of every cultural question those models help us ask).

\section{Measuring structure}\label{sec:structure}

The first type of cultural measurement involves imposing a formal structure on multi-party dialogue.
The agential cut here is \textit{structural}: any conversational exchange can be formalized in multiple ways---as a sequence of turns, a tree of reply-to links, a network of topic threads---and each formalization draws a different boundary between what counts as ``structure'' and what is relegated to noise. 

Conversation analysis has long studied the systematics of turn-taking~\citep{Sacks1974-qs} and the collaborative work of speakers and hearers~\citep{Goodwin1981-zs}; choosing among formalizations is an interpretive commitment about which of those systematics the apparatus will be allowed to see.
Choosing reply-to graphs makes conversational floor, address, and initiation visible at scale; it forecloses lexical cohesion, topical drift, and the slow buildup of mutual understanding that fluent dialogue depends on.
The gender measurements that follow are visible only inside this cut---they would not survive a re-formalization that, for example, weighted topic continuity over reply structure.

\subsection{Conversation disentanglement}

The structural formalization I adopt here is \textit{conversation disentanglement}: recovering the thread structure of interleaved multi-party dialogue, a task studied extensively in NLP on IRC chat logs~\citep{Elsner2008-ni,Kummerfeld2019-hj,Jiang2018-dr,Zhu2021-cc}.
In \citet{chang-etal-2023-dramatic}, we extend this to scripted multi-party dialogue, developing a BERT-based model~\citep{devlin-etal-2019-bert} that predicts which prior utterance each line responds to, thereby recovering a latent thread structure.
Formally, given a sequence of utterances $\{u_1, \ldots, u_n\}$, the model encodes each utterance contextually and scores candidate reply-to links:
\begin{equation}
    P(\text{parent}(u_i) = u_j) \propto \exp\big(g(\boldsymbol{h}_i, \boldsymbol{h}_j)\big),
\end{equation}
where $\boldsymbol{h}_i$ is the contextual representation of utterance $u_i$ and $g$ is a learned scoring function.
The result is a directed graph---a thread structure---extracted from continuous dialogue.

In my framework, this is an apparatus-dependent measurement: the threads do not pre-exist in the script but are produced by the apparatus.
Here, the start of a thread is grounded in observations in television studies:~\citet{McKee2016-tt} argues that speech acts are driven by character need: ``all talk responds to a need, engages a purpose, and performs an action.''
This is central to our annotation scheme, which itself is part of the apparatus: what counts as a ``reply,'' whether it breaks a thread or continues it---these are not neutral transcription decisions but interpretive commitments that shape the thread graph the model produces.
At the same time, the original work includes exhaustive experiments across architectures and input representations---different encoders, different context windows, different representations of speaker identity---each constituting a different apparatus configuration applied to the same dialogue.
These are design choices that, in addition to enabling comparison between methods, reflect what we take a speaker to be, which in turn shapes the ensuing analysis.
Each combination of annotation scheme, encoder, and input representation enacts a different agential cut on the same underlying dialogue---and the cultural patterns that emerge are inseparable from the apparatus that produced them.

\subsection{Measurements of agency}

\begin{figure}
  \centering
  \includegraphics[width=0.9\columnwidth]{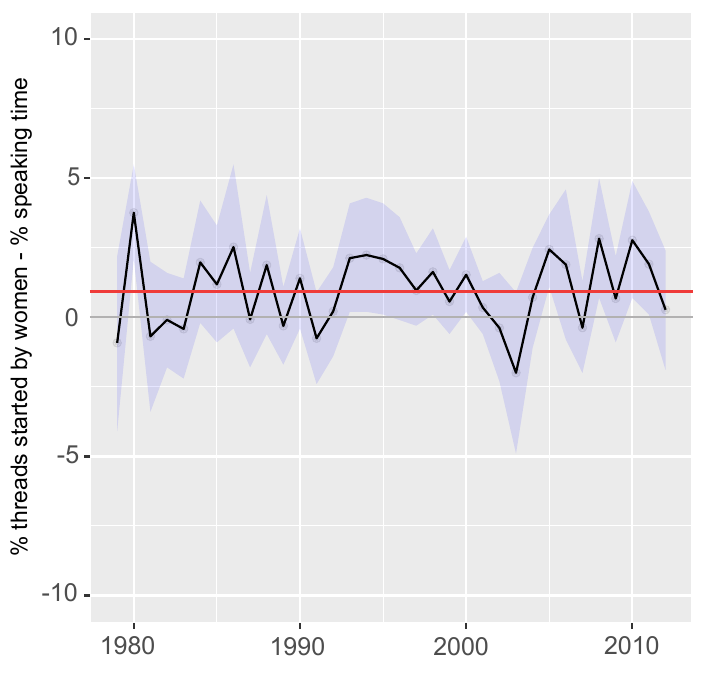}
	\caption{Percentage of conversational threads started by female characters minus their share of speaking time, by year, with 95\% confidence intervals (shaded;~\citealp{chang-etal-2023-dramatic}). The red line marks the overall average of $+1.0$ percentage points ($p < 0.05$): despite their under-representation, women initiate threads at a rate that slightly exceeds their speaking time.}
  \label{fig:gender-threads}
\end{figure}

Applied at scale to TV and movie transcripts, the resulting thread structures enable measurements of conversational agency: who initiates threads, who sustains them, and how these patterns distribute by gender.
In our data from 808 movies, 30.4\% of threads are started by female characters---consistent with the well-documented disparity in women's presence and voice in media~\citep{Rosen1973-cq,Hooks_bell1992-zj}.
However, when normalized by each character's share of speaking time (Fig.~\ref{fig:gender-threads}), women initiate threads at a rate that slightly exceeds their speaking time, with an average absolute difference of $+1.0$ percentage points ($p < 0.05$).

This finding is surprising: despite their under-representation, female characters are written to claim the conversational floor more often than their male counterparts would predict.
This project demonstrates what structural measurement---an affordance of NLP applied to cultural objects---makes possible: by committing to a particular formalization of conversational structure, the apparatus renders visible patterns of agency that would be invisible in the unstructured transcript.

\section{Measuring interaction}\label{sec:interaction}

The study of how language use varies with interactional context has deep roots: \citet{Ervin-Tripp1964-rs} showed that speakers shift register depending on topic, setting, and listener; \citet{Ng1993-na} documented how verbal behavior both reflects and constitutes power asymmetries; and studies of scripted dialogue have long recognized that television writing encodes---and sometimes contests---these dynamics~\citep{Richardson2010-kg,Bednarek2023-zu}.

To address such cultural questions, the second type of cultural measurement involves classifying participants into sociolinguistic categories---speaker, addressee, side-participant---that formalize who participates and how.
The agential cut here is \textit{categorical and modal}: the apparatus partitions a continuous audio-visual stream into a finite set of discrete role labels, and the choice of which roles to recognize---and which modalities to admit as evidence---determines what aspects of interactional positioning can register at all.

\subsection{Conversational role attribution}

Multimodal video understanding has produced large-scale datasets for television---most notably TVQA~\citep{lei2018tvqa}, which benchmarks compositional question answering over TV clips.
But existing datasets treat dialogue as a source of answers rather than as an interactional system with its own structure.
To address this gap, we operationalize the sociolinguistic organization of conversation itself to devise an annotation scheme, culminating in TV-MMPC~\citep{chang2026multimodalconversationstructureunderstanding}.

The annotation scheme is itself an agential cut: Goffman's theorization of conversation participants, along with \citeposs{clark1982hearers} taxonomy of speakers and hearers, is mapped onto discrete labels assignable to individual utterances in scripted television, so the model is tasked, for each utterance, to predict its speaker, intended addressee, and side-participants.
This cut turns the continuous, multimodal flow of conversation into a discrete assignment of roles, and the choice of role categories is itself a discursive commitment about which aspects of interactional positioning matter enough to measure.

In this particular case, we evaluate a range of models: a text-only model discards everything non-verbal; a multimodal model takes in the full audio-visual signal and maps it into the same label space.
These are different apparatuses measuring the same phenomenon, and they produce different results---not only in model performance, but in what counts as a relevant signal: which modalities the researcher selects, which frames the vision--language model samples, whether audio and video are processed jointly or separately, and what computing resources make feasible.

\subsection{Measurements of gendered roles}

{
\newlength{\imgboxheight}
\setlength{\imgboxheight}{4.5cm} 

\begin{figure*}[t]
  \centering
  \begin{subfigure}[t]{0.342\textwidth}
    \centering
    \subcaption{}
    \begin{minipage}[t][\imgboxheight][t]{\linewidth}
      \includegraphics[width=\linewidth,keepaspectratio]{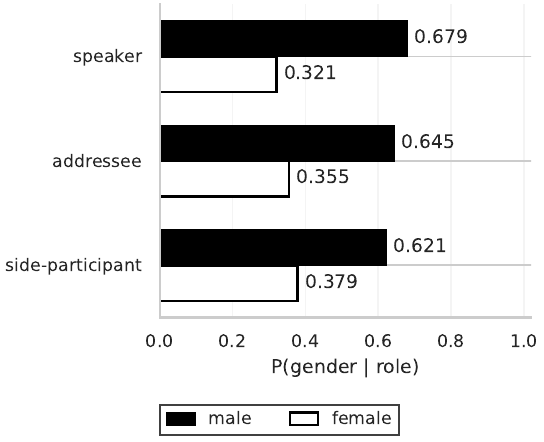}
    \end{minipage}
  \end{subfigure}\quad%
  \begin{subfigure}[t]{0.272\textwidth}
    \centering
    \subcaption{}
    \begin{minipage}[t][\imgboxheight][t]{\linewidth}
      \includegraphics[width=\linewidth,keepaspectratio]{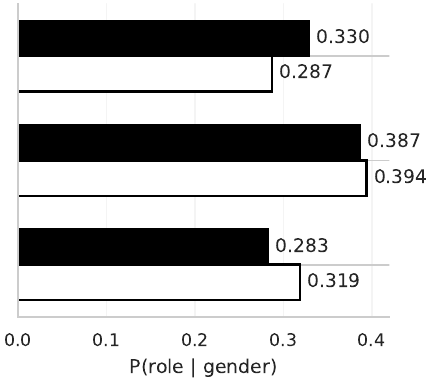}
    \end{minipage}
  \end{subfigure}\quad%
  \begin{subfigure}[t]{0.143\textwidth}
    \centering
    \subcaption{}
    \begin{minipage}[t][\imgboxheight][t]{\linewidth}
      \includegraphics[width=\linewidth,keepaspectratio]{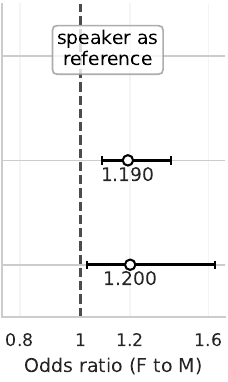}
    \end{minipage}
  \end{subfigure}
  \caption{
 Relationship between gender and conversational roles~\cite{chang2026multimodalconversationstructureunderstanding}. 
  }
  \label{fig:three-panels-tablelike}
\end{figure*}
}

Applied to 350,842 utterances across four TV series in TVQA, the apparatus produces measurements of how interactional roles distribute by gender.
Fig.~\ref{fig:three-panels-tablelike} breaks down this relationship. 
Panel (a), showing $P(\text{gender}|\text{role})$, confirms that men are the numerical majority in every role. 
However, Panel (b), which shows $P(\text{role}|\text{gender})$, reveals a structural disparity:
men are more likely than women to occupy the \textit{speaker} role (33.0\% of their roles vs.\ 28.7\% for women).
Conversely, this pattern inverts for listening roles: women are more likely than men to be cast as an \textit{addressee} (39.4\% vs.\ 38.7\%) and a \textit{side-participant} (31.9\% vs.\ 28.3\%).
This skewed distribution positions male characters as primary speakers and female characters more frequently as the audience.
Panel (c) formalizes this disparity: we fit a multinomial logistic regression that estimates the log-odds of occupying each role, controlling for show-level effects.
With speaker as the reference category and $j \in \{\text{addressee, side-participant}\}$:

{\small\begin{align}
\log \frac{P(\text{role}=j)}{P(\text{role}=\text{speaker})} = \beta_{j,0} &+ \beta_{j,\text{female}} \cdot \mathbb{I}(\text{female}) \notag\\
 & + \sum_{s=1}^{S-1} \gamma_{j,s} \cdot \mathbb{I}(\text{show}_s),    
\end{align}}

\noindent This is itself a measurement: from the high-dimensional space of individual role attributions to a two-dimensional summary (one odds ratio per non-speaker role) that isolates the effect of gender.
The resulting odds ratios---$\exp(\beta_{j,\text{f}})$---are 1.19 for addressee and 1.20 for side-participant (both $p<0.001$), indicating that women are approximately 1.2 times as likely as men to appear in a listening role rather than as speaker, after controlling for show.

These measurements provide evidence of how social roles and power dynamics are constructed---and reinforced---in cultural artifacts.
They are visible only because the apparatus includes the categories of addressee and side-participant: the theoretical commitment to distinguishing listening roles is what makes the gendered pattern measurable.

\section{Measuring deviation}\label{sec:deviation}

The third type of cultural measurement involves a normative baseline and quantifying departures from it.
Where structural measurement asks ``what is the form?'' and interaction measurement asks ``who participates how?'', deviation measurement asks ``where does practice depart from expectation?''
The agential cut here is \textit{normative}: the apparatus learns a baseline of stereotypic patterns from training data, and what counts as ``subversion'' exists only relative to that learned norm. A different training corpus would yield a different baseline; a different label inventory---one that included \texttt{queer-platonic} or \texttt{enmeshed sibling}---would yield different deviations.
What this cut makes measurable is stereotypicality and its breach; what it forecloses is meaning that neither stabilizes as a stereotype nor disrupts one---the in-between cases that read as simply ordinary.

Deviation, then, is a relational property: between texts, and between text and apparatus.
This case study takes inspiration from cognitive stylistics, what \citet{Culpeper2001-go} calls ``stereotyping'': the textual cues by which dramatic figures are constructed as social beings provide the signal, and the model's trained expectations provide the norm against which deviation becomes measurable.

\subsection{Stereotypic relation extraction}

In \citet{chang2024subversive}, we train a Longformer~\citep{Beltagy2020-vx} encoder on 787 digitized pilot teleplays to predict the social relationship enacted in a dyad's dialogue.
Given a scene $\mathcal{S}$ with utterances from a head character $c_h$ and tail character $c_t$, the model builds a joint representation.
A Longformer encoder extracts the CLS token for each speaker's concatenated utterances.
To incorporate scene context beyond the target speakers' words, an attentive pooling mechanism weights the hidden states of the full scene, guided by a token-level mask $M$ that zeros out the target speakers' tokens:
\begin{align}
    \alpha &= \mathrm{softmax}\big({\boldsymbol{w}_A}^\top \boldsymbol{h}_{\mathcal{S}} \odot M\big), \label{eq:attn}\\
    \boldsymbol{h} &= \big[\boldsymbol{e}_{\texttt{<s>}}^{c_h};\; \boldsymbol{e}_{\texttt{<s>}}^{c_t};\; {\boldsymbol{h}_{\mathcal{S}}}^\top \alpha\big], \label{eq:rep}
\end{align}
where $M[j]=0$ if token $j$ is spoken by either target speaker and $1$ otherwise, $\boldsymbol{h}_{\mathcal{S}}$ is the encoded scene, and $\boldsymbol{w}_A$ is a learned attention vector.
The concatenated representation $\boldsymbol{h}$ is projected through a linear classification head: 
\begin{equation}
p(r|\boldsymbol{h}) = \mathrm{softmax}(f(\boldsymbol{h})),	
\end{equation}
predicting a relationship type $r$ from a fixed set (e.g., \texttt{colleague\_of}, \texttt{sibling\_of}, \texttt{spouse\_of}).

The architecture enacts a specific agential cut.
The mask $M$ directs the model to attend to what \textit{other} characters say and do in the scene---the ambient social context---rather than relying solely on the target dyad's words.
This is a deliberate choice about what the apparatus should treat as signal: the broader conversational ecology, not just the dyad in isolation.
Character names are anonymized to force the apparatus to operate on dialogic cues---\textit{how} characters talk---rather than on memorized character-relationship associations (see \S\ref{sec:practice} for the empirical consequences of this choice).

\subsection{Measurements of subversion}

In traditional NLP, model quality is measured by accuracy: how well predictions match ground-truth labels.
This evaluation paradigm treats correct labels as the goal and errors as failures to be minimized.
But for cultural measurement, this logic inverts---a lesson that working at the intersection of NLP and cultural analysis has made unavoidable.
Metrics like accuracy are measurements of \textit{performance}; what cultural analysis requires are measurements of \textit{interpretive significance}.

The model in \citet{chang2024subversive} is deliberately trained as a ``stereotyping reader'': it learns what sibling dialogue, or spouse dialogue, \textit{typically} sounds like across hundreds of teleplays.
Rather than evaluating accuracy as an end in itself, the work measures \textit{subversion} as the discrepancy between the model's predicted distribution over relationship types and the ground-truth labels.
When a model trained on stereotypical patterns of sibling dialogue predicts that two brothers sound like a married couple, the ``error'' is the finding.
The gap between prediction and reality is the very signal to be analyzed---a measurement of how interaction departs from the norm the apparatus has learned.

\begin{figure}[t]
  \centering
  \includegraphics[width=0.85\columnwidth]{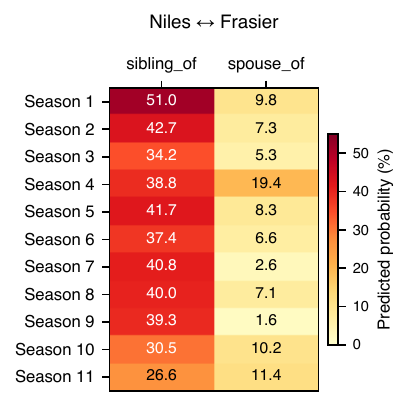}
  \caption{Percentage of predictions for \texttt{sibling\_of} (ground truth) and \texttt{spouse\_of} between Niles and Frasier Crane across \textit{Frasier}~\citep{chang2024subversive}.}
  \label{fig:frasier}
\end{figure}

Applied to \textit{Frasier}, the apparatus identifies the Crane brothers' dialogue as consistently deviating from stereotypical sibling interaction.
The model frequently predicts \texttt{spouse\_of} for Frasier and Niles's exchanges---capturing, and formalizing, what queer theorists would recognize as a form of intimacy that exceeds its nominal category~\citep{Sedgwick2003-wq,Halperin2002-ym}: the ``crypto-gay''~\citep{Clum1999-tg} quality that cultural critics have noted in their bickering, codependent, linguistically ornate relationship (Fig.~\ref{fig:frasier}).
The apparatus produces a formal trace of the gap between performed relationship and normative type that can then become an object of cultural analysis.
\citeposs{Butler1990-rb} notion of the ``subversion of identity'' here becomes computationally tractable: notably, it is not meant to be an ontological claim about the characters' sexuality, but as a measurable discrepancy between dialogic performance and the model's trained expectations.

\section{Between measuring instruments and cultural artifacts}\label{sec:practice}

The three case studies above focus on what NLP can measure; in this section, we turn to the apparatus itself, through three examinations: \textit{erasure}, removing the apparatus's likely purchase on its memory of the cultural artifact in question to reveal what was being measured (\S\ref{sec:erasure}); \textit{attunement}, deliberately adapting a model on a corpus it has likely not seen to see what measurement looks like when the apparatus may have little memory of its object to draw on (\S\ref{sec:attunement}); and \textit{agency}, distributing the measurement across an agentic workflow of several backbones, to see how the boundary between instrument and object shifts (\S\ref{sec:agency}).
Together they describe the apparatus from more than one side---what it brings to a culturally familiar text, what it has to build for a culturally distant one, and what it inherits from the other models it is chained to---and show where the boundary between the measuring instrument and the cultural artifact it measures is actually being drawn.

\subsection{Erasure}\label{sec:erasure}

To see what a measurement actually rests on, we can take away what the model may already remember and watch what happens to the result.
In the multimodal conversation structure study~(\S\ref{sec:interaction}), replacing character names with anonymous identifiers (e.g., renaming \texttt{Sheldon Cooper} to \texttt{Character C}) collapses speaker recognition from 78.6 to 13.7 and addressee recognition from 68.1 to 15.7 (Table~\ref{tab:anon}, a).
The same model, on the same data, appears to rely on its parametric memory for role attribution; what appeared to be a boundary between a neutral model and a conversational structure was, it seems, a boundary between two layers of the model's own cultural knowledge: its general language competence and its specific memory of these characters, invoked by their names.
Anonymization forces a different cut, one that separates dialogic structure from character identity, and the performance collapse reveals which cut was operative all along.

\begin{table}[t]%
    \centering
    \small
    \setlength{\tabcolsep}{3.5pt}
	\begin{tabular}{@{}llccc@{}}
    \toprule
     & \textbf{Task} & \textbf{Orig.} & \textbf{Anon.} & $\boldsymbol{\Delta}$ \\
    \midrule
    \multicolumn{5}{@{}l}{(a) Multimodal conversation structure} \\
    \midrule 
    & \textbf{Gemini 2.0 Flash} & & & \\
    & \quad Speaker (Acc) & 78.6 & 13.7 & $-$64.9 \\
    & \quad Addressee (F$_1$) & 68.1 & 15.7 & $-$52.5 \\
    \midrule
    \multicolumn{5}{@{}l}{(b) Stereotypic relation extraction} \\
    \midrule
    & \textbf{Longformer} & & & \\
    & \quad Role (Acc) & 34.8 & 24.8 & $-$10.0 \\
    & \quad + anon.\ training & 36.7 & 33.8 & $-$2.9 \\
    & \textbf{LLaMA 3-70b} & & & \\
    & \quad Role (Acc) & 24.3 & 19.7 & $-$4.6 \\
    \bottomrule
    \end{tabular}

    \caption{Effects of anonymization across two studies on different test sets: original (``Orig.'') and anonymized (``Anon.''). Panel (a): for multimodal conversation structure~\citep{chang2026multimodalconversationstructureunderstanding}, speaker and addressee recognition collapse under anonymization; panel (b): for stereotypic relation extraction~\citep{chang2024subversive}: the scene attentive pooling model without anonymized training shows a 10-point accuracy gap; training on anonymized data recovers most of the performance.}
    \label{tab:anon}
\end{table}

The same pattern recurs in stereotypic relation extraction (\S\ref{sec:deviation}; Table~\ref{tab:anon}, b).
The Longformer model trained only on unanonymized data drops 10 points when evaluated on anonymized test data (from 34.8 to 24.8), indicating that the model exploits character names to infer memorized relationships rather than parsing the dialogue.
Training on anonymized data recovers most of this gap, which forces the model to attend to how characters talk rather than who they are.
LLaMA 3-70b~\cite{Grattafiori2024-uz} zero-shot drops from 24.3 to 19.7 under anonymization, a smaller gap than Gemini's but the same pattern nonetheless.
Prompt-based LLMs lack the Longformer's most direct mitigation (adaptation with anonymized data), though input-side anonymization can offer partial alternatives.

These anonymization experiments contextualize the cultural findings from the case studies: the measurements of gendered role attribution and relational subversion are shaped by the model's prior cultural formation.
When Gemini attributes a speaking role or LLaMA predicts a character relation, the output reflects the cultural material internalized during training as much as the signal in the input.
The material---what the model was trained on, what cultural knowledge its parameters encode---and the discursive---which categories the researcher defines, which tasks the model performs---are entangled in every measurement.
This entanglement is at the heart of using language models as measurement apparatus for culture: the instrument might have already internalized some of the cultural material it measures, and the research narrative must account for it, rather than treat data, model, and evaluation as clearly separable stages of a pipeline.

\subsection{Attunement}\label{sec:attunement}

The anonymization analysis showed what the apparatus does when the cultural material may be available in training data.
The inverse case is more revealing about what the apparatus \textit{is} when there is no shortcut through prior knowledge, which I explore here in the context of Restoration comedy (1660--1700).
Restoration drama is interesting here because it is well-studied in history and criticism but computationally under-explored, in part because the professionally curated and digitized editions sit behind proprietary access. 
The language is historically distant, and an off-the-shelf LLM, especially a smaller one, has at best a thin grasp of its conventions.

Restoration comedy of manners is driven by characters and archetypes (such as \textit{rake} and \textit{fop}). 
That said, we do not know \textit{a priori} how many lines we need to have read for a character's archetype to become recognizable, which resembles a sufficient-context problem~\citep{lopez-monroy-etal-2018-early} in which the reader does not recognize an archetype all at once but accumulates evidence as the play unfolds, committing when predictions have stabilized enough to act. 
This problem lets us specify two dimensions of the apparatus directly: what its parameters have been tuned to, and how it reads.

The corpus for this toy experiment is a random sample of 109 plays (1,283 character episodes) from Chadwyck-Healey English Drama collection;\footnote{See Appendix~\ref{app:data} for more details.} each episode $e = (x_{1:T}, y)$ is an ordered sequence of one character's lines.
Formally, a backbone $f$ encodes prefix $x_{1:t}$ into a representation $\boldsymbol{h}_t = f_{\theta}(x_{1:t})$, from which a classifier predicts an archetype label $c$:
\begin{equation}
p_{\theta}(c \mid x_{1:t}) = \mathrm{softmax}\big(\boldsymbol{W}\boldsymbol{h}_t\big)_c, \quad c \in \mathcal{C}.
\label{eq:restoration_classifier}
\end{equation}
Drawing on taxonomies defined in \citet{hirst-1979-comedy,mast-1975-comic}, we include the following in $\mathcal{C}$: \textsc{Rake}, \textsc{Fop}, \textsc{Natural}, \textsc{Obstacle}, to be predicted from a character's dialogue alone, absent any metadata. 
The source data has no archetype labels; we developed them iteratively---reading Restoration criticism, hand-annotating, refining prompts---and then scaled with Gemini 2.5 Flash~\cite{Gemini-Team2023-oj}, achieving Cohen's $\kappa = 0.71$ against the author's annotation of five plays.

To test this apparatus, we compare two reader-models on the same 1B-parameter Gemma backbone~\citep{Gemma-Team2024-no}---a fixed-window reader (first $N$ lines) and the entropy-thresholding reader together with a majority-class baseline.
Entropy thresholding models the epistemic experience of the reader: the apparatus reads sequentially, stops when predictive entropy falls below $\delta$, and predicts the maximum-probability class:
\begin{align}
H_t &= -\!\!\sum_{c \in \mathcal{Y}} p_{\theta}(c \mid x_{1:t}) \log p_{\theta}(c \mid x_{1:t}), \label{eq:entropy} \\
\tau_{\delta}(e) &= \min\{t : H_t \leq \delta\}, \label{eq:tau} \\
\hat{y}(e) &= \arg\max_{c} p_{\theta}(c \mid x_{1:\tau(e)}). \label{eq:pred}
\end{align}
For evaluation, we track macro-F$_1$ alongside an efficiency-aware objective $\mathcal{J} = \text{macro-F}_1 - \lambda \cdot \overline{\rho}$ ($\lambda = 0.25$ by default, $\rho(e) = \tau(e)/T$ denotes the fraction of lines consumed), which prices each unit of reading against the prediction it enables.

Attunement comes through one epoch of continued pre-training on roughly 174,000 lines of in-domain dialogue, in the spirit of historically attuned LMs~\citep{manjavacas-arevalo-fonteyn-2021-macberth}.
The effect is consistent across reader-models: $\mathcal{J}$ rises from 0.17 to 0.29 for fixed-window and from 0.25 to 0.37 for entropy thresholding.
The attuned entropy-thresholding reader reaches macro-F$_1$ 0.44 reading only 26\% of lines, against the fixed-window reader's 0.38 at 36\% and a majority baseline of 0.14.
Attunement gives the apparatus a thinner, deliberately-built version of the cultural past; the apparatus itself is layered: a human-calibrated annotator-LLM defines the norm against which an attuned reader-LLM is measured.

Taken together, \S\ref{sec:erasure} and \S\ref{sec:attunement} examine the same instrument from two angles.
In the first, the apparatus's prior cultural formation is already there and can be perturbed; in the second, it has to be built up, and even built carefully it cannot reach beyond the cuts that defined it.
In both cases, the measuring instrument and the cultural artifact are entangled in every result; neither produces it alone.

\subsection{Agency}\label{sec:agency}

Erasure and attunement each treat a single model as the measuring instrument, whose parametric knowledge could be perturbed or built up.
One model may not always be sufficient, however.
Measuring anything about a two-hour film, for instance, strains that picture, because the film often cannot enter a model whole: 
context and file-size limits cap most models, and most models do not take video natively at all, so the raw film must be processed first (e.g., frames sampled, audio transcribed, scenes segmented).
Each reduction is a cut that fixes what is even available to be measured, and the agentic turn in AI multiplies these cuts: a workflow may caption the video with one model, reason over the captions with another, and retrieve what seems relevant with a third, so agency appears to move into an autonomous system.

We see this in our benchmark designed to measure how well LLMs can understand classical Hollywood films~\cite{Bamman2026-by}, which is challenging because many narrative phenomena are long-range (e.g., spanning across scenes) and multimodal (requiring both audio and visual input).
At the core of evaluation are different ways of compressing the video of a film for video QA models: watching a sampled fraction of the frames, reading an ASR transcript, reasoning over generated captions, or retrieving frames on demand.
Other than feeding the entire movie into a natively multimodal LLM like Gemini, we find it effective to turn the film into a chronological sequence of clip-based captions that a separate backbone (Claude or GPT) reasons over; the answering model never sees the original video.

What's surprising here is that, given the same captions, Claude is largely indifferent to whether captions arrive as full context or through agentic retrieval, while Gemini performs significantly worse in the latter setup.
In other words, the measurement moves when the same artifact is routed through a different model position, and Claude-as-retriever is not Claude-in-context.
Here, no model is the measuring instrument on its own; the so-called agentic workflow and the video compression pipeline that interact with the answering backbone form an apparatus in perhaps the most obvious way.%

Memory complicates that boundary further, albeit less obviously.
Popular films are widely discussed online, so a model can answer from metapragmatic knowledge \textit{about} a film rather than its content: prompted with only a title and release year, three frontier models answer roughly 40\% of the questions correctly (Gemini 39.4, Claude Opus 40.7, GPT 38.8), well above the 25\% chance rate, so we retain only the 628 of 779 questions that none can answer from metadata alone.
This filter is not a cleanup step before the measurement but another cut: a constitutive exclusion deciding what counts as measuring the film rather than recalling it.
At the same time, we also observe memorization alone does not drive reasoning performance: per-film accuracy correlates \textit{negatively} with a film's online prevalence and with how well a model can name it from its frames, and though Gemini recognizes these films far more often than the others (28.4\% against 2--3\%), that recognition yields no answering advantage.
This also surfaces an contrast between Claude and Gemini: the former works more naturally with agents, and the latter carries more knowledge about classical Hollywood in itself.

As part of a measurement apparatus, an LLM is never merely handed tools to orchestrate; it occupies a position---captioner, retriever, answerer---and, as a result, becomes a site where the cultural artifact is constituted for measurement.
What looks like an interaction among LLM-based agents, in this light, is better read as \textit{intra-action}~\citep{barad2007}: where interaction assumes separate entities that exist first and then meet, intra-action holds that the parties come to be distinct only through their relation---boundaries and properties are produced in the encounter, not brought to it.
Agency here is not the autonomy the agentic turn ascribes to a single system; it is enacted by the apparatus: the cut between the culture an instrument has absorbed and the culture set before it as object.

\section{Conclusion}\label{sec:conclusion}

In this paper, I have argued that NLP work on cultural objects is a material-discursive practice in which the apparatus participates in producing the phenomena it measures. 
Every measurement here is \textit{culturally contingent}: the findings depend on the specific apparatus through which they were produced.
What sets the measurement of culture apart from other forms of task and evaluation is that the instrument itself is culturally situated: it carries the training distribution's biases, its era's textual archive, its architecture's affordances.
The concepts of agential cut and material-discursive practice provide a framework for taking this reflexive entanglement seriously.

This framework has implications for how measurements of culture can be taken and interpreted:
accuracy and model performance remain necessary, but they are insufficient.
The anonymization experiments, for instance, show that what a model gets \textit{wrong}---evidence of memorization, deviation from stereotypical expectation---can be more productive than what it gets right.
More fundamentally, optimizing for a task and interrogating the contingency of the measurement are complementary: the former establishes what the apparatus can do, the latter reveals what the apparatus is made of---and neither can be separated from the situated practice that produces it.

Cultural analytics is best understood as an interdisciplinary experiment that treats computational work as a material-discursive practice: it ceaselessly reflects on---and attempts to redefine---its positionality between positivist tradition and the negative movement of theory, while interrogating the reciprocal relations among humans, data, and information that undergird it, in order to shed new light on the worlds we inhabit.
To measure culture with a language model, then, is to turn the instrument on itself, which requires that agential cuts be conscious methodological and ethical commitments---and learning to do so deliberately, rigorously, and productively completes the big picture.

\section*{Acknowledgments}

I thank my thesis advisor, Prof. David Bamman, for years of guidance on the work synthesized here, and my collaborators on the projects underlying this paper, named in the original publications, for the partnership that made each contribution possible.
I am also grateful to the Stanford Literary Lab for sharing the Chadwyck-Healey English Drama source data used in \S\ref{sec:attunement}, and to Prof. Mark Algee-Hewitt and Emil Wang for helpful discussions.
I thank the anonymous reviewer for their feedback.

Most of all: the students of my Cultural Analytics class at UC Berkeley in Fall 2025, subtitled ``Machine Learning and Measurements of Culture.''\footnote{\url{https://ca.kentkc.org}}
The ideas represented here emerged, died, transformed over the semester that we spent together.
Teaching and learning with you was among the most generative experiences of my time at Berkeley.
Thank you for helping me see the big picture.

\bibliography{custom}

\appendix

\section{Restoration Drama Corpus}\label{app:data}

The corpus underlying \S\ref{sec:attunement} draws on the Chadwyck-Healey English Drama collection, hand-transcribed by the original publisher with a bespoke XML schema and shared by the Stanford Literary Lab. 
The raw transcripts are processed through a two-pass agentic pipeline. In the first pass, Gemini reads each source file with an annotated view of its page-break tags and produces a structured manifest via a Pydantic schema: the manifest segments the file into plays by title, author, and page range, and recovers a cast map linking canonical character names to abbreviated speech tags. 
In the second pass, a LangGraph state machine walks each play page-by-page, maintaining the active play, the current speaker, and a short context window of previous lines; for each page, the model classifies lines as \texttt{SPEECH}, \texttt{STAGE\_DIRECTION}, \texttt{ACT\_HEADER}, \texttt{SCENE\_HEADER}, \texttt{PROLOGUE}, \texttt{EPILOGUE}, or \texttt{PARATEXT}, normalizes speaker names against the cast map, and detects transitions between plays in multi-play source files.

The pipeline ran over 166 source files, producing 598 play-level segments. 
From these, 113 TSVs were retained for downstream analysis, 111 of which were readable, and 109 yielded at least one valid character episode under our preprocessing rules. 
The author first annotated five randomly sampled plays, which informed the design of the pipeline described above, and then used Gemini 2.5 Flash to generate the character-episode archetype labels upstream.
Cohen's $\kappa$ between the Gemini and human labels was 0.71; broader human annotation remains future work.

An \textit{episode} is the dialogue produced by one character in one play, kept in dramatic order, retained only if it contains $\geq 3$ lines and has a majority archetype label among the four classes. 
Across 109 plays we obtain 1,283 episodes (median 49 lines per episode, maximum 1,110). 
Data is split by play, not by character, yielding 82 train, 11 development (for hyperparameter selection), 16 test plays, and 972, 116, 195 episodes, respectively. 

The continued pre-training data is constructed by concatenating each play's speaker-attributed lines in dramatic order, totaling 174,016 lines across 111 play-level documents.
Continued pre-training uses causal next-token modeling on \texttt{google/gemma-3-1b-pt} for one epoch with block size 1024, learning rate $2 \times 10^{-4}$, per-device batch size 1, and gradient accumulation 16, without LoRA or 4-bit quantization. 
Because this stage uses unlabeled dialogue from the corpus as a whole, including held-out plays, it should be understood as unsupervised domain adaptation.

\end{document}